\documentclass[10pt,reqno]{amsart}       
\usepackage{graphicx}
\usepackage{amsmath,amssymb,enumerate,caption}
\newtheorem{theorem}{Theorem}

\begin{document}
\title{Optimal Release Time Decision from Fuzzy Mathematical Programming Perspective }
\maketitle

\begin{center}
Arvind Kumar$^1$, Adarsh Anand$^2$, Pankaj Kumar Garg$^3$, Mohini Agarwal$^4$

\vspace{5mm}
 $^1$Department of Mathematics, University of Delhi, Delhi 110007, India\\
              {arvind.ch83@gmail.com}, {akumar1@maths.du.ac.in}\\
\vspace{3mm}
 $^{2, 4}$Department of Operational Research, University of Delhi, Delhi 110007, India\\
              {adarsh.anand86@gmail.com}, {mohini15oct@gmail.com}\\
\vspace{3mm}
 $^3$Department of Mathematics, Rajdhani College, University of Delhi, Delhi 110015, India\\
              {gargpk08@gmail.com}
\vspace{3mm}
\end{center}

\begin{abstract}
Demand for high software reliability requires rigorous testing followed by requirement of  robust modeling techniques for software quality prediction. On one side, firms have to steadily manage the reliability by testing it vigorously, the optimal release time determination is their biggest concern. In past many models have been developed and much research has been devoted towards assessment of release time of software. However, majority of the work deals in crisp study. This paper addresses the problem of release time prediction using fuzzy Logic. Here we have formulated a Fuzzy release time problem considering the cost of testing under the impact of warranty period. Results show that fuzzy model has good adaptability.
\end{abstract}

\section*{1. Introduction}
Software development life cycle $(SDLC)$ consists of five stages namely: Analysis, Design, Coding, Testing and Operational phase. Testing phase of $SDLC$ plays a very important role in determining the quality of the software. The main objective of testing phase is to remove as many faults as possible with a minimum cost. The three main quality attributes are viz Reliability, Scheduled Delivery and Cost and the primary objective of the software developer's is to attain them at their best values. On the other hand software user's requirements conflicts with the developers. Software users demand faster deliveries, cheaper software and quality product, whereas software developers aim at minimizing their development cost, maximizing the profit margins and meeting the competitive requirements. The resulting situations call for a tradeoff between conflicting objectives of software user's requirement with the developers. As a course of best alternative the developer management must determine optimally when to stop testing and release the software system to the user focusing on users requirements, simultaneously satisfying their own objectives. Such a problem is known as software release time decision $(SRTD)$ problem ~\cite{13},~\cite{14}.

On time release of the software benefit developers in two ways: Firstly they can obtain the maximum returns on their investments, secondly they can satisfy the conflicting users requirement. Premature release of the software may require a lot of time to fix the faults after the release and can suffer from goodwill loss. Also the delay in software release impose penalty cost, revenue loss etc. Thus one must optimally determine the release time of the software to reduce the losses that can be imposed due to early or later release of the software ~\cite{13},~\cite{14}. Such a problem of software reliability discipline can be formulated as an optimization problem with single or multiple objectives under some well-defined sets of constraints. There are a number of crisp and soft computing methodologies, optimization techniques and routines to solve such problems.

Making use of traditional $SRGM$, many optimization problems have been formulated in literature. These models have helped to determine the relationship between the testing progress and time. Okumoto and Goel ~\cite{22} derived simplest release time policies based on exponential $SRGM$ in two ways. In the first approach they considered an unconstrained cost objective while in the other, they considered the unconstrained reliability objective. The problem was formulated assuming all the costs are deterministic and well defined, as well as the level reliability required to achieve is determined on the basis of experience by the management. Later other researchers followed the approach with different consideration and improvements.

Yamada and Osaki ~\cite{25} discussed release time problems with cost minimization objective under reliability aspiration constraint and reliability maximization objective under constraint based on exponential, modified exponential and S-shaped $SRGM$. Kapur and Garg ~\cite{21} discussed the concept of considering penalty cost by introducing the concept of releasing the software at scheduled delivery time set by the management and$ \backslash $or with an agreement between the user and developer on release time problem. Kapur and Garg ~\cite{19} discussed release policies using exponential, modified exponential and S-shaped test effort based $SRGM$ for maximizing expected gain function subject to achieving a given level of failure intensity. Huang ~\cite{4} and Huang and Lyu ~\cite{5} discussed release policies considering the effect of testing effort expenditure.

Yun and Bai ~\cite{26} proposed that software release time problems should assume software life cycle to be random as several factors such as availability of alternative competitive product in the market, a better announcement by the developer himself, etc. Later Kapur et al ~\cite{17} determined release for a software system based on minimizing expected cost subject to achieve a desired level of intensity assuming software life cycle to be random. Further, Kapur et al ~\cite{18} developed a software cost model incorporating the cost of dependent faults along with independent faults. Pham and Zhang ~\cite{23} modified the traditional cost function by incorporating warrant and risk cost.

Kapur et al ~\cite{20} developed a simple cost model to include separate cost of fixing a fault due to perfect and imperfect fault debugging during testing and operational phase along with the testing cost per unit time. Pham ~\cite{24} discussed a release policy for a fault complexity based pure fault generation $SRGM$. Along with the above mentioned basic cost model they included the penalty cost in the cost function and defined the operational life cycle length to be random. Further, Kapur and Garg ~\cite{20}, Huang ~\cite{4} and Huang et al. ~\cite{6} have considered the effect of fault removal efficiency in determining the release time of the software. Later on some researchers work on bi-criterion release time problem, Such as Kapur et al. ~\cite{16} developed a multi-objective optimization problem for determination of release time.  In which the considered two simultaneous objective functions as reliability maximization and cost minimization for release time determination, they assigned weights to the two objective functions according to their relative importance. Owing to the importance of change point concept Kapur et al. ~\cite{13} formulated a release policy for the exponential change point $SRGM$ in which they modified the simple cost model as given by Okumoto and Goel ~\cite{22} to include the cost of fault removal before and after the change point.

Most of the problems on release time even up to the recent times have been formulated using static conditions. Crisp optimization techniques such as method of calculus, Lagrange multipliers or crisp mathematical programming techniques were used to solve the problem. Crisp modeling has complications such as ~\cite{1} :

1.	Real situation are not very often crisp and deterministic and they cannot be prescribed precisely.

2.	The complete description of a real system often would require more detailed data than a human being could ever recognize simultaneously, process it and understand.

Crisp means yes-or-no type rather than more-or-less type. Many management problems involve decision making under the ambiguous information. Due to conditions prevailing in the market and competitive reasons, the developers can only make ambiguous statements on the organizations goals and available resources bringing uncertainty in the problem definition. There are various other sources that bring uncertainty in the computation such as system complexity, subject's awareness, communication and thinking about uncertainty, intended flexibility, complex relationships between the various variables and economics of information. One widely accepted solution of this problem is to define the problem under fuzzy environment, as it offers the opportunity to model subjective imagination of the decision maker. Fuzzy set theory offers a precise mathematical form to describe fuzzy terms in the form of linguistic variables. To represent the shades of meaning of such linguistic terms, the concept of possibility values of membership has been introduced. Membership of an object will vary like ~\cite{1}:

\begin{itemize}
  \item[$\bullet$]{For no membership}
  \item[$\bullet$]{For full membership}
  \item[$\bullet$] {For partial membership}
\end{itemize}

Fuzzy set theory builds a model to represent a subjective computation of possible effect of the given values on the problem and permits the incorporation of vagueness in the conventional set theory that can be used to deal with uncertainty quantitatively. Fuzzy optimization is a flexible approach that permits a more adequate solution of real problems in the presence of vague information, providing the well defined mechanisms to quantify the uncertainties directly. It has proven to be an efficient tool for the treatment of fuzzy problems. Another advantage of fuzzy set theory is that it saves lot of time required for enormous information processing in order to determine average values in the classical modeling due to its capability to directly operate on vague information.

Kapur et al ~\cite{12} formulated a software release time decision-making problem with cost minimization goal subject to failure intensity constraint under fuzzy environment and discussed the solution methodology with numerical illustrations. Jha, Indumati and Kapur ~\cite{7} formulated a discrete $SRGM$ based single criteria constrained release policy for software system under fuzzy environment. Jha et al. ~\cite{8} proposed the fuzzy bi-criterion constrained release time problem based on discrete $SRGM$ by using two conflicting fuzzifier objectives of cost minimization and reliability maximization simultaneously subject to the crisp budgetary constraint. Jha et al. ~\cite{9} formulated a fuzzy bi-criterion constrained release time problem by using two objectives of cost minimization and reliability maximization simultaneously subject to a failure intensity constraint on two types of Imperfect debugging and Error Generation $SRGM$ under Fuzzy environment. Jha, Indumati and Kapur ~\cite{10} developed an interactive mathematical programming approach for a bi- criterion unconstrained release time problem for a flexible $SRGM$ under two stage Yamada model ~\cite{13} with two types of imperfect debugging and logistic fault removal model.

In this paper, an optimal release time of the software is determined based on minimizing the cost and achieving the desired level of reliability under fuzzy environment. Further, we discussed the approach to solve fuzzy optimization problem by transforming it into a crisp problem. The resulting problem may be feasible or infeasible. If the crisp problem is feasible then the optimal solution is also the solution of fuzzy problem but if it is infeasible the approach of goal programming can be applied to obtain a compromised solution of the problem.

Rest of the paper is organized as follows: Section-2 gives a brief knowledge about the $SRGM$ used in the cost modeling. Modeling of the cost structure and problem formulation is discussed in section-3 while in the next section some basic terminologies of fuzzy theory and the solution procedure is presented in section-5 describing in detail about the fuzzy optimization technique. Numerical analysis of the above proposed model is given in section-6. Finally conclusion and references are presented.

\section*{2. Software Reliability Growth Model}

To formulate the release time problem for minimizing the cost incurred during testing and debugging under the warranty period; we used the cost structure as given by Pham  and Zhang ~\cite{23} for determining the optimal release time of the software. In this paper, we use the $SRGM$ given by Goel and Okumoto ~\cite{3} to describe the expected failure phenomenon. The following assumptions are used:
Assumption
\begin{enumerate}
\item[1.]	{Failure observation / Fault removal phenomenon is modeled by $NHPP$.}
\item[2.]	{Software is subject to failures during execution caused by faults remaining in the software.}
\item[3.]	{Each time a failure is observed immediate efforts takes place to find the cause of the failure in order to remove it.}
\item[4.]	{Failures are observed during execution caused by remaining faults in the software.}
\end{enumerate}

Following differential equation describes the failure phenomenon of the Goel and Okumoto model ~\cite{3} based on above set of assumptions.
 \begin{equation}\label{eq:1}
 \frac{d}{dt}m(t)=b(a-m(t))
 \end{equation} 									
Solving equation (1) under the initial condition $m(0)=0$,  we get the following mean value function
\begin{equation}\label{eq:2}
  	m(t)=a(1-e^{-bt})	
\end{equation}								
And the reliability   of the software is given as follows:
  								
\begin{equation}\label{eq:3}
  	R((T+T_W)|T)= e^{-(m(T+T_W)-m(T))}	
\end{equation}	
\section*{3. Problem Formulation}							
In this section we discuss an optimal release time problem based on the methodology to minimize the total expected cost testing subject to a reliability level to be achieved till the time of release of the software. Here, we have considered the cost structure given by Pham and Zhang ~\cite{23} to describe the total expected cost. Accordingly we have, $C_0$ to be the fixed set up cost of testing the software. Under the assumption that it takes time to remove faults and removal time of each fault follows a truncated exponential distribution the probability density function of the time to remove a fault during testing period $Y$, can be given by,
\begin{align}
 s(y)=
 \begin{cases}
 \frac{\lambda_{y} e^{- \lambda_{y} y}}{\int_0^{T_{0}} \lambda_{y} e^{- \lambda_{y} x}  \mathrm{d}x} &\quad \textrm{for $0\leq y\leq T_{0}$} \\
 0 &\quad \textrm{for $y>T_{0}$}
 \end{cases}
 \end{align}
   where $\lambda_{y}$  is a constant parameter associated with truncated exponential density function $Y$  and $T_{0}$  is the maximum time to remove any error during testing period. Then the expected time to remove each error is given by 	

\begin{equation}\label{eq:5}
  	E(Y)= \mu_{y}=\int_0^{T_{0}} y .s(y)  \mathrm{d}y = \frac{1-(\lambda_{y}{T_{0}}+1)e^{-\lambda_{y}{T_{0}}}}{ \lambda_{y}(1-e^{-\lambda_{y}{T_{0}}})}
 \end{equation}
Hence, the expected total time to remove $N(T)$ faults corresponding to all failures experienced up to time $T$ is given by

\begin{equation}\label{eq:6}
  	E\bigg(\sum_{i=1}^{N(T)}{Y_{i}}\bigg)= E(N(T)).E(Y_{i})= m(T). \mu_{y}
 \end{equation}
 Thus, the expected cost to remove all errors detected by time $T$ in the testing phase, where $C_{1}$ is now the cost of removing an error per unit time during testing phase is given by

 \begin{equation}\label{eq:7}
  	E_{1}(T)=C_{1}E\bigg(\sum_{i=1}^{N(T)}{Y_{i}}\bigg)= C_{1} m(T). \mu_{y}
 \end{equation}
The cost of testing per unit time is assumed to be a power function of time $T$ since the cost of testing increases with higher gradient in the beginning and slow down later.

\begin{equation}\label{eq:8}
  	E_{2}(T)= C_{2}T^{\alpha}
 \end{equation}
 Further it is assumed that the software developer does not maintain the software for the whole of its operational life cycle. This is because the software developers always keep on improving their software and come up with newer versions with added features and improved reliability. The newer versions are usually launched even before the earlier version obsoletes and the developers encourage the users of the previous versions to improve their version with the new one as it has enhanced features. So, given any version of the software, developers decides a warranty period for which they provide after sales services and after that period if a failure is encountered no removal is made from the part of the developer. Hence now instead of calculating the cost for the whole life cycle of the software we need to calculate it only up to the time when the warranty period ends. The expected cost to remove all faults during warranty period $[T, T+T_{w}]$ is given by
 \begin{equation}\label{eq:9}
  	E_{3}(T)= C_{3}\mu_{w}(m(T+T_{w})-m(T))
 \end{equation}
 Therefore, the total expected cost can be summed up to:

 \begin{align}\label{eq:10}
  	E[C(T)] &= C_{0}+E_{1}[T]+E_{2}[T]+E_{3}[T]\\
  &=C_{0}+C_{1}m(T)\mu_{y}+C_{2}T^{\alpha}+C_{3}\mu_{w}(m(T+T_{w})-m(T))
  \end{align}
The optimal release time problem can be formulated as follows:
 \begin{align*}
 &\text{Min $C(T)$}\\
&\text{Subject to $R(x|T)\tilde{\geq} R_{0}$ \quad \quad \quad \quad \quad \quad \quad \quad \quad \quad \quad \quad \quad \quad \quad \quad \quad \quad \quad \quad \quad \quad \quad \quad \quad \quad \quad \quad \quad \quad  $(P1)$}\\
\end{align*}
where the inequality in the reliability constraint is not precise. Infact the management may be ready to accept a level of reliability lesser than $R_{0}$. The symbol is called fuzzy greater than or equal to and have linguistic interpretation i.e. essentially greater than or equal to. The crisp optimization problems can be solved by the mathematical programming approach. If the desired level of reliability by the time is too high and the available resources are very limited then the problem becomes infeasible and can be solved by the goal programming approach to find a compromised solution. Solution obtained using goal programming is sensitive to the waiting vectors and goals of the problem. Crisp optimization techniques have no well-defined mechanisms to handle the uncertainties in the problem formulation. Hence fuzzy optimization techniques (Fuzzy Mathematical Programming, Fuzzy Goal Programming, etc) are used to solve the fuzzy optimization problems. In the following section we discussed some basic terminologies of fuzzy optimization techniques as available in literature ~\cite{1}.

\section*{4. Basic Definition of Fuzzy Set Theory and Fuzzy Number}

In this section we introduce some of the basic terminologies of fuzzy set theory and present various set theoretic operations ~\cite{1}.

\begin{enumerate}
\item	\textbf{Fuzzy Set}: Let $X$ be the universe whose generic element is denoted by $x$ . A fuzzy set $A$ in $X$ is a function $A:X\rightarrow[0,1]$
\item	\textbf{Support of a Fuzzy Set}: Let $A$ be a fuzzy set in $X$. Then the support of $A$, denoted by $S(A)$, is the crisp given by $S(A)=\{x\in X:\mu_{A}(x)>0\}$
\item	\textbf{Standard Union}: The standard union of two fuzzy sets $A$ and $B$ is a fuzzy set $C$ whose membership function is given by $\mu_{C}=\text{max}\{\mu_{A}(x),\mu_{B}(x)\} \text{for all} x \in X$   This we express as $C=A\cup B$
\item	\textbf{Standard Intersection}: The standard intersection of two fuzzy sets $A$ and $B$ is a fuzzy set $C$ whose membership function is given by $\mu_{C}=\text{min}\{\mu_{A}(x),\mu_{B}(x)\} \text{for all} x \in X$   This we express as $C=A\cap B$
\item	\textbf{$\alpha-$cut} : Let $A$ be a fuzzy set in $X$ and $\alpha (0,1]$ The $\alpha-$cut of the fuzzy set $A$ is the crisp set $A_{\alpha}$ given by $A_{\alpha =\{x\in X:\mu_{A}(x)>\alpha \}}$
\item	\textbf{Convex Fuzzy Set}: A fuzzy set in $R^n$  is called a fuzzy set if its $\alpha-$cut $A_{\alpha}$  are (crisp) bounded sets for all $\alpha \in (0,1].$
\item	\textbf{Fuzzy Number}: A fuzzy set in $R^n$ is called a fuzzy number if it satisfies the following conditions:
\begin{enumerate}
\item[i.]{$A$ is normal}
\item[ii.]{$A$ is convex}
\item[iii.] { $\mu_{A}$ is upper semi continuous}
\item[iv.]	{The support of $A$ is bounded}
\end{enumerate}
\end{enumerate}
The theorem presented below gives a complete characterization of a fuzzy number.
\begin{theorem}
Let $A$ be a fuzzy set in $R$. Then $A$ is a fuzzy number if and only if there exists a closed interval (which may be a singleton) $[a,b]\neq \Phi$ such that
\begin{align*}
 \mu_{A}(x)=
 \begin{cases}
 1, &\quad x\in [a,b]\\
 l(x),&\quad x\in (-\infty,a)\\
 r(x), &\quad x \in (b, \infty)
 \end{cases}
 \end{align*}
 Where
 \begin{enumerate}
 \item[i.]{$l:(-\infty,a)\rightarrow [0,1]$   is non-decreasing, continuous from the right and $l(x)=0$  for $x\in (-\infty,w_{1}),w_{1}< a$}
\item[ii.] {$r:(b,\infty)\rightarrow [0,1]$ is non-increasing, continuous from the left and $r(x)=0$  for $x\in (w_{2},\infty), w_{2}> b$
and $\mu(x)$  is called Membership Function of Fuzzy Set A on R.}
\end{enumerate}
\end{theorem}
For solving the fuzzy mathematical problems Zimmermann's approach can be used. In this approach first the objective function of the fuzzifier minimization is understood in the sense of the satisfaction of a restriction level $C_{B}$  and appropriate membership functions are defined for fuzzy inequalities and then employ Bellman and Zadeh ~\cite{2} principle to identify the fuzzy decision, which results in a crisp mathematical programming problem algorithm.

In the next section we have used the Zimmermann's approach to solve the Fuzzy Optimization Problem.
\section*{5. Solution of the Release Time Problem: Fuzzy Optimization Technique}
Fuzzy Optimization Zimmermann's ~\cite{27} approach is discussed for solving the Fuzzy Release Time Problem formulated in section-$3$. The first step is to restate the problem with fuzzifier in objective function included as a restriction level constraint. The problem can be restated as follows:
\begin{align*}
\text{Find $\quad \quad T$}\\
\text{Subject to  } &C(T)< C_{B}\\
 & R(X|T)>R_{0}\\
& T\geq 0 \text{\quad \quad \quad \quad \quad \quad \quad \quad \quad \quad \quad \quad \quad \quad \quad \quad \quad \quad \quad \quad \quad \quad \quad \quad \quad \quad \quad \quad \quad \quad \quad \quad \quad  $(P2)$}
\end{align*}
We define the membership function $\mu_{i}(T); i=1,2$ for each of the fuzzy inequality in problem

\begin{align*}
 \mu_{1}(T)=
 \begin{cases}
 1  &\quad ;C(T)\leq C_{0}\\
 \frac {C^*-C(T)}{C^*-C_{0}} &\quad ;C_{0}<C(T)\leq C^*\\
 0  &\quad ;C(T)\geq C^*
 \end{cases}
 \end{align*}
\begin{align*}
 \mu_{2}(T)=
 \begin{cases}
 1  &\quad ;R(x|T)\geq R_{0}\\
 \frac {R(x|T)-R^*}{R_{0}-R^*} &\quad ;R^*<R(x|T)\leq R_{0}\\
 0  &\quad ;R(x|T)\leq R^*
 \end{cases}
 \end{align*}
 where $C^*$ and $R^*$ are the respective tolerance for the reliability level and available resources. Next we use Bellman and Zadeh's ~\cite{2} principle to identify the fuzzy decision to solve the fuzzy system of inequality corresponding to the problem. The resulting crisp optimization problem is given as

\begin{align*}
&\text{Maximize} \quad \quad  \alpha \\
&\text{Subject to  }  \mu_{i}(T)\geq \alpha \quad i=1, 2; \quad \alpha \geq 0, \quad T \geq 0 \text{\quad \quad \quad \quad \quad \quad \quad \quad \quad \quad \quad \quad \quad \quad \quad \quad \quad \quad \quad  $(P3)$}
 \end{align*}
If the problem is feasible then an optimal solution can be found using crisp mathematical programming approach else if the problem is infeasible then we use crisp goal programming approach to solve the problem and identify the fuzzy goal decisions. The problem P3 after incorporating parameter values can be solved by mathematical programming approach using $LINGO$ or any other software package.
\section*{6. Numerical Example}
For application of the above formulated release policy let us consider the data set reported by Musa et al. based on the failure from real time command and control system. Using this data set the parameters of Goel Okumoto ~\cite{3} are estimated to be $a= 143.32$  and $b=0.1246$ . Let us suppose that cost parameters are  $C_{0}=\$ 50, \quad C_{1}=\$ 60, \quad C_{2}=\$ 700$  and  $C_{4}=\$ 3600$. If the operational mission time  $x=1$ CPU hour, the warranty period length is  $T_{w}= 450$ CPU hour, $\alpha = 0.95, \mu_{w}= 0.5$  and $\mu_{y}=0.1$ . The release time problem based on following data can be analyzed. Further total budget $C_{\beta}=\$ 26000$   and the reliability requirement by the release time is $R_{0}= 0.95$  with tolerance on cost and reliability $C^*=\$ 31000$  and $R^*= 0.80$ (we have assumed these values for illustration, however these values are set by management based on past experience). Using values of various parameters and constants, solution of the problem is obtained with fuzzy optimization method discussed above. The mean value function for failure phenomenon and reliability function with estimated values of the parameters is given below:

\begin{align*}
m(T)= 143.32(1-e^{-0.1246T})\text{$\quad$and$\quad$} m(T+T_{w})= 143.32(1-e^{-0.1246(T+450)})\\
 R((T+T_{w})|T)= e^{-(m(T+450)-m(T))}\text{$\quad$and$\quad$  } R(x|T)= e^{-(m(T+450)-m(T)}
 \end{align*}
The membership function for the fuzzy cost and reliability constraint are given as

$\mu_{1}(T)= \frac {(31000-(50+60*143.32*(1-e^{-0.1246T})*0.1+700*T^{0.95}+3600*0.5*143.32*(e^{-0.1246T}-e^{-0.1246(T+450)})))}{(31000-26000)}$\\

where $26000 < C(T)\leq 31000$

$\mu_{2}(T)= \dfrac {e^{-(m(T+1)-m(T))}-0.8}{0.95-0.80}\quad \text{where} \quad 0.80 \leq e^{-(m(T+1)-m(T))}< 0.95$
\begin{figure}\centering\includegraphics[height=5cm,width=13cm]{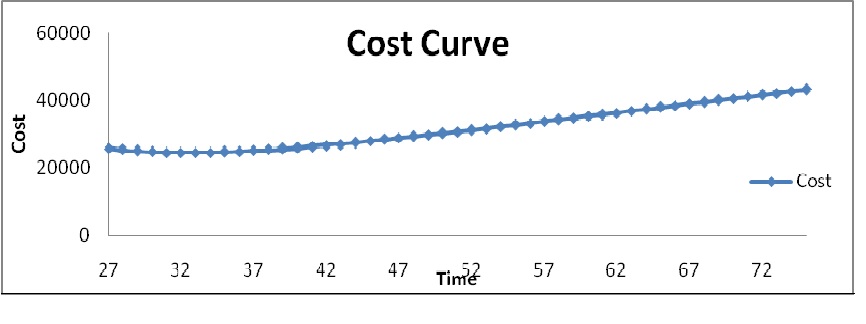}\caption{ Cost Curve}\label{fig:Stupendous}\end{figure}
\begin{figure}\centering\includegraphics[height=5cm,width=13cm]{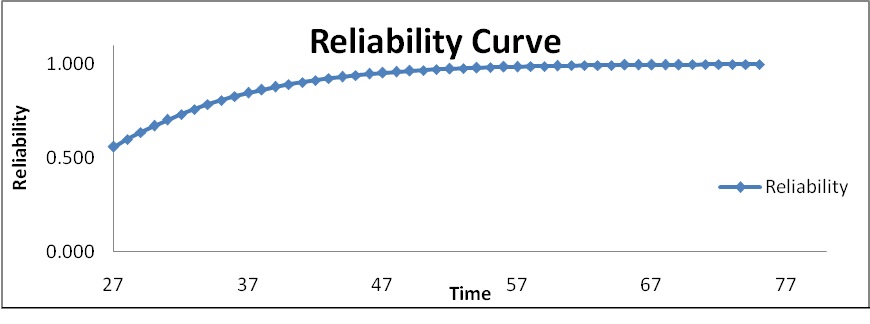}\caption{Reliability Curve}\label{fig:Stupendous}\end{figure}
Figure-1 shows the monotonically increasing trend of cost curve over time. The curve of reliability over time is shown in figure- 2. The cost and reliability membership functions plotted on cost and reliability scales respectively are shown in figure 3 and 4.

\begin{figure}\begin{center}\includegraphics[height=5.2cm,width=13cm]{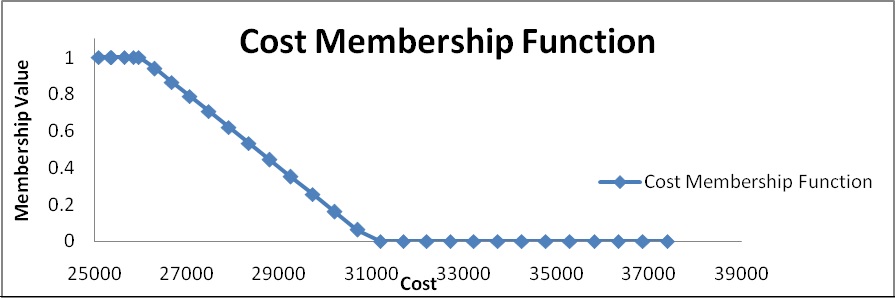}\caption{Cost Membership Function.\label{fig:Stupendous}}\end{center}\end{figure}

\begin{figure}\begin{center}\includegraphics[height=6cm,width=13cm]{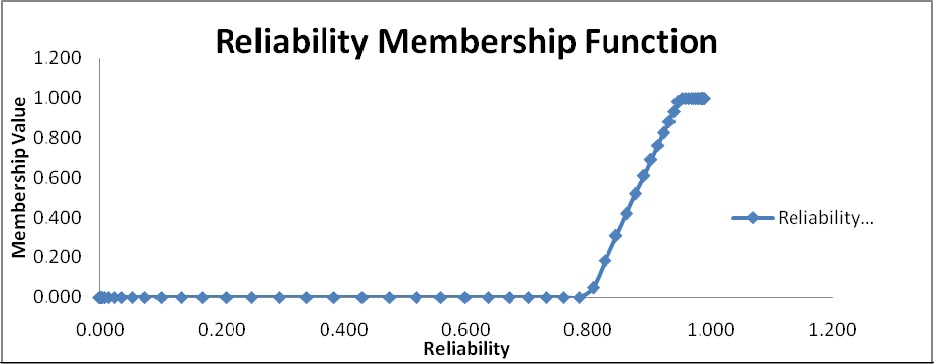}\caption{Reliability Membership Function.\label{fig:Stupendous}}\end{center}\end{figure}

The two membership functions can now be plotted simultaneously on time scale as shown in figure 5. Figure 5 shows that the two curves intersect at a point, which gives the optimal release time of the software and can be found by solving the crisp mathematical programming problem using standard mathematical programming approach in $LINGO$ software.
\begin{figure}\begin{center}\includegraphics[height=6cm,width=13cm]{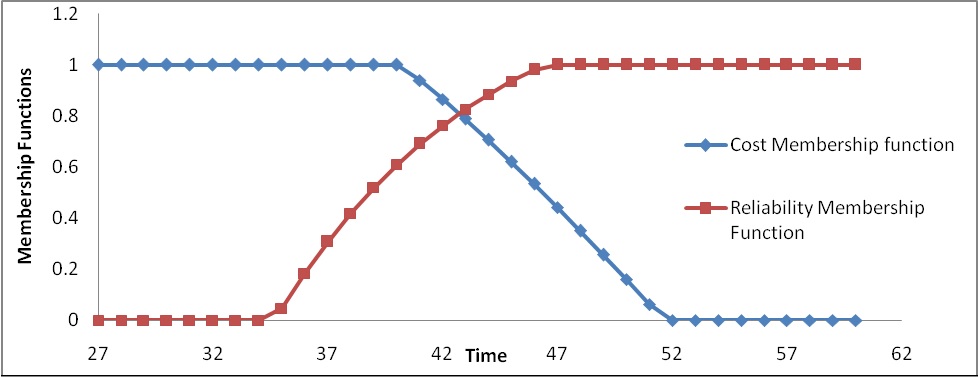}\caption{Membership Function of Cost and Reliability (Feasible) .\label{fig:Stupendous}}\end{center}\end{figure}

\begin{align*}
&\text{ Maximize $ \alpha$ \quad \quad \quad \quad \quad \quad \quad \quad \quad \quad \quad \quad \quad \quad \quad \quad \quad \quad \quad \quad \quad \quad \quad \quad \quad \quad \quad \quad \quad \quad \quad \quad \quad \quad \quad \quad \quad \quad \quad \quad \quad \quad \quad \quad \quad \quad \quad \quad \quad \quad \quad \quad \quad \quad \quad \quad \quad \quad \quad \quad \quad \quad \quad \quad \quad \quad}\\
&\text{ Subject to \quad \quad \quad \quad \quad \quad \quad \quad \quad \quad \quad \quad \quad \quad \quad \quad \quad \quad \quad \quad \quad \quad \quad \quad \quad \quad \quad \quad \quad \quad \quad \quad \quad \quad \quad \quad \quad \quad \quad \quad \quad \quad \quad \quad \quad \quad \quad \quad \quad \quad \quad \quad \quad \quad \quad \quad \quad \quad \quad \quad \quad \quad \quad \quad \quad \quad}\\
\end{align*}
$\mu_{1}(T)= \frac {(31000-(50+60*143.32*(1-e^{-0.1246T})*0.1+700*T^{0.95}+3600*0.5*143.32*(e^{-0.1246T}-e^{-0.1246(T+450)})))}{(31000-26000)}\geq \alpha$\\
\begin{align*}
&\mu_{2}(T)= \frac {e^{-(m(T+450)-m(T))}-0.8}{0.95-0.80}\geq \alpha \quad \quad \quad \quad \quad \quad \quad \quad \quad \quad \quad \quad \quad \quad \quad \quad \quad \quad \quad \quad \quad \quad \quad \quad \quad \quad \quad \quad \quad \quad \quad \quad \quad\\
& \alpha \geq 0, \quad \alpha \leq 1, \quad T\geq 0\quad \quad \quad \quad \quad \quad \quad \quad \quad \quad \quad \quad \quad \quad \quad \quad \quad \quad \quad \quad \quad \quad \quad \quad \quad \quad \quad \quad \quad \quad \quad \quad \quad
\end{align*}

Solving the problem (P5) we obtain optimal release time $T^*=42.72$  and $\alpha^*= 0.809$  and the cost  $C(T^*)= 26949.769$ and  $R(x|T^*)= 0.9213$ .

If suppose the available amount of total testing resources is changed from  $C_{B}=26000$  to $C_{B}=23000$  with tolerance level changed from $C^*=31000$  to $C^*=24500$  then we see that coat and reliability function don't intersect at any point as shown in figure 6 which indicate that the crisp problem is infeasible.

\begin{figure}[htb]\begin{center}\includegraphics[height=6cm,width=13cm]{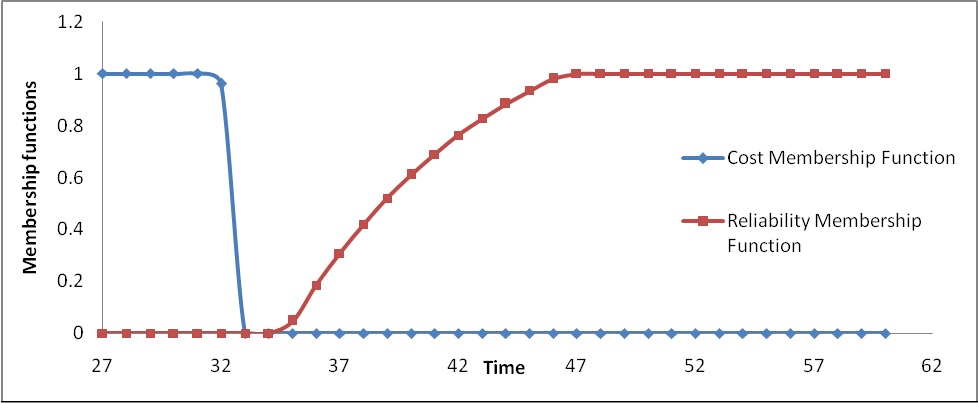}\caption{Membership Function of Cost and Reliability (Infeasible) .\label{fig:Stupendous}}\end{center}\end{figure}

In such a situation we solve the problem using the Goal Programming approach to obtain a fuzzy goal solution. In Goal Programming approach, management aspirations and restrictions are achieved by minimizing both positive and negative deviations. The problem is restated as follows:
\begin{align*}
\text{Maximize} \quad \quad  &\eta_{1}+\eta_{2} \\
\text{Subject to $\quad $ } & \mu_{1}+ \eta_{1}-\rho_{1}=\alpha\\
&\mu_{2}+ \eta_{2}-\rho_{2}=\alpha\\
 &\alpha \geq 0, \quad \alpha \leq 1, \quad T \geq 0\\
 &\eta, \rho \geq 0 \quad ;i=1,2 \text{\quad \quad \quad \quad \quad \quad \quad \quad \quad \quad \quad \quad \quad \quad \quad \quad \quad \quad \quad \quad \quad \quad \quad \quad \quad  $(P4)$}
\end{align*}

Solving the above problem by LINGO software we obtain the optimal release time $T^*=34.68$   and $\eta_{1}=0.105$ . At the optimal release time the cost is 24657.35 which is 157.35 more than the maximum tolerance level and reliability is 0.80.
\section*{7. Conclusion}
In this paper, we have formulated a fuzzy release time problem and discussed the fuzzy mathematical programming solution procedure with an illustration when a feasible solution exist and if the problem is infeasible fuzzy goal programming approach is used to solve the problem. The global market has become fiercely competitive and the software developers have many competitive goals to be optimized simultaneously. Hence the release time of the software should be determined simultaneously optimizing the various conflicting objectives.
\section*{8. Acknowledgments} Research of Arvind Kumar is supported by University Grants Commission, New Delhi, India", Sch. No./JRF/AA/283/2011-12. The authors are grateful to Ratnesh R. Saxena for valuable suggestions which helped in improving the paper.


\begin{thebibliography}{110}
\bibitem{1}Bector C.R., Chandra S., \emph{Fuzzy Mathematical Programming and Fuzzy Matrix Games,} Springer-Verlag Berlin Heidelberg, (2005).
\bibitem{2}Bellman R.E., Zadeh L.A., \emph{Decision making in fuzzy environment}, Management Science, (1973), 17, 141-164
\bibitem{3}Goel, A.L., Okumoto, K., \emph{Time dependent error detection rate model for software reliability and other performance measures}, IEEE Transaction Reliability, R-28(3), (1979), 206-211.
\bibitem{4}Huang C. Y., \emph{Cost-reliability-optimal release policy for software reliability models incorporating improvements in testing efficiency}, Journal of system Software, 77, (2005), 139-155.
\bibitem{5}Huang C. Y. and Lyu, M. R., \emph{Optimal Release Time for Software Systems considering Cost, Testing-Effort and Test Efficiency}, IEEE transactions on Reliability, 54(4), (2005), 583-591.
\bibitem{6}Huang C. Y., Kuo, S.Y. and Lyu, M. R., \emph{ Optimal Software Release Policy based on Cost and Reliability with  Testing  Efficiency},  In: Proceedings of 23rd IEEE annual international computer software and applications conference, Phoenix, AZ, (1999) 468-473.
\bibitem{7}Jha, P. C., Indumati and Kapur P. K., \emph{Optimal Release Policy for a Discrete Flexible Model incorporating the effect of Fault Removal Efficiency under Fuzzy Environment}, P. C. Jha and M. N. Hoda (Ed.), Mathematical Modeling, Optimization and Their Applications, Narosa Publishing house, New Delhi, (2011), 7-19.
\bibitem{8}Jha, P. C., Indumati, Singh O. and Gupta, D. \emph{ Bi-criterion Release Time Problem for a Discrete SRGM under Fuzzy Environment}, International Journal of Mathematics in Operational Research, 3(6), (2011), 680-696.
\bibitem{9}Jha, P. C., Singh O., Indumati and Kapur, P.K., \emph{Bi-criterion Release Time Problem Incorporating Effect of Two types of Imperfect Debugging under Fuzzy Environment}, Om Parkash (Ed.), Advances in Information Theory and Operations Research: Interdisciplinary Trends, VDM Verlag, Germany, (2010), 142-153.
\bibitem{10}Jha, P. C., Indumati and Kapur, P.K., \emph{Interactive Approach to Release Time Problem of Software under Fuzzy Environment}, Communication in Dependability and Quality Management, an International Journal, 3(3), (2010), 61-75.
\bibitem{11}Jha, P. C., Kumar, D. and Kapur, P.K., \emph{ Fuzzy Release Time Problem}, International Conference on Quality, Reliability and Infocom Technology (ICQRIT-06), MacMillan India Ltd, (2007), 304-310.
\bibitem{12}Kapur, P.K., Pham, H., Gupta, A. and Jha, P. C., \emph{Optimal release policy under fuzzy environment}, International Journal of Systems Assurance Engineering and Management, 2(1), (2011), 48-58.
\bibitem{13}Kapur P.K., Pham H., Gupta A., Jha P.C., \emph{Software Reliability assessment with OR application}, Springer, Berlin, (2011).
\bibitem{14}Kapur P.K., Garg R.B., Kumar S., \emph{Contribution to hardware and software reliability}, World Scientific publishing Co. Pvt. Ltd., Singapore, (1999)
\bibitem{15}Kapur, P.K., Garg, R.B, Aggarwal A.G. and Tandon, A., \emph{ General Framework for Change Point problem in Software Reliability and Related Release Time Problem}, In: Proceedings ICQRIT, (2009).
\bibitem{16}Kapur, P.K., Agarwal, S. and Garg, R. B., \emph{Bi-criterion Release Policy for Exponential Software Reliability Growth Models}, Recherche Operationanelle / Operational Research, 28, (1994), 165-180.
\bibitem{17}Kapur, P.K., Garg, R. B. and Bhalla, V. K., \emph{Release Policy with Random Software Life Cycle and Penalty Cost}, Microelectron Reliability, 33(1), (1993), 7-12
\bibitem{18}Kapur, P.K. and Garg, R.B., \emph{A Software Reliability Growth Model for an Error Removal Phenomenon}, Software Reliability Journal, (1992), 291-294.
\bibitem{19}Kapur, P.K. and Garg, R. B., \emph{Optimal Software Release policies for Software Systems with testing Effort}, International Journal System Science, 22(9), (1991), 1563-1571.
\bibitem{20}Kapur, P.K. and Garg, R. B., \emph{Optimal Software Release policies for Software reliability Growth models under Imperfect Debugging}, Recherche Operationanelle / Operational Research, 24, (1990), 295-305.
\bibitem{21}Kapur, P.K. and Garg, R. B., \emph{Cost-Reliability Optimum Release Policies for Software System under Penalty Cost}, International Journal System Science, 20, (1989), 2547-2562.
\bibitem{22}Okumoto, K. and Goel, A. L., \emph{Optimum Release Time for Software Systems Based on Reliability and Coat Criteria}, Journal of system Software, 1, (1980), 315-318
\bibitem{23}Pham H., Zhang X., \emph{A software cost model with warranty and risk costs}, IEEE Trans Comp 48(1),  (1999), 71-75
\bibitem{24}Pham, H., \emph{A Software Cost Model with Imperfect Debugging, Random Life Cycle and Penalty Cost}, International Journal System Science, 27, (1996), 455-463.
\bibitem{25}Yamada, S. and Osaki, S., \emph{Optimal Software Release Policies with simultaneous Cost and Reliability Requirements}, European Journal of Operational Research, 31, (1987), 46-51.
\bibitem{26}Yun, W.Y. and Bai, D.S., \emph{Optimum Software Release Policy with Random Life Cycle}, IEEE transactions on Reliability, 39(2), (1990), 338-353.
\bibitem{27}Zimmermann, H.J., \emph{Fuzzy set theory and its applications}, Academic Publisher (1991).

\end{thebibliography}
\end{document}